\pdfoutput=1
\documentclass[runningheads]{llncs}
\usepackage{graphicx}
\usepackage{amsmath}
\usepackage{multirow}
\usepackage{multicol}
\usepackage{makecell}
\usepackage{amssymb}
\usepackage{siunitx}
\usepackage{tabularx}
\usepackage[misc]{ifsym}
\begin{document}
\title{Superpixel Cost Volume Excitation for Stereo Matching\thanks{Supported by National Natural Science Foundation of China (Grant No.
41927805). 
}}


\author{Shanglong Liu\ \and
Lin Qi{(\textrm{\Letter})} \and
Junyu Dong{(\textrm{\Letter})} \and
Wenxiang Gu \and
Liyi Xu}

\authorrunning{S. Liu et al.}

\institute{School of Computer Science and Technology, Ocean University of China,  Qingdao 266101, China\\
\email{\{liushanglong, gwc1323, xly3385\}@stu.ouc.edu.cn}\\
\email{\{qilin,dongjunyu\}@ouc.edu.cn}}

\maketitle              

\begin{abstract}
In this work, we concentrate on exciting the intrinsic local consistency of stereo matching through the incorporation of superpixel soft constraints, with the objective of mitigating inaccuracies at the boundaries of predicted disparity maps.
Our approach capitalizes on the observation that neighboring pixels are predisposed to belong to the same object and exhibit closely similar intensities within the probability volume of superpixels. By incorporating this insight, our method encourages the network to generate consistent probability distributions of disparity within each superpixel, aiming to improve the overall accuracy and coherence of predicted disparity maps.
Experimental evaluations on widely-used datasets validate the efficacy of our proposed approach, demonstrating its ability to assist cost volume-based matching networks in restoring competitive performance.
\end{abstract}
\begin{figure}[t]
  \centering
  \centerline{\includegraphics[width=\columnwidth]{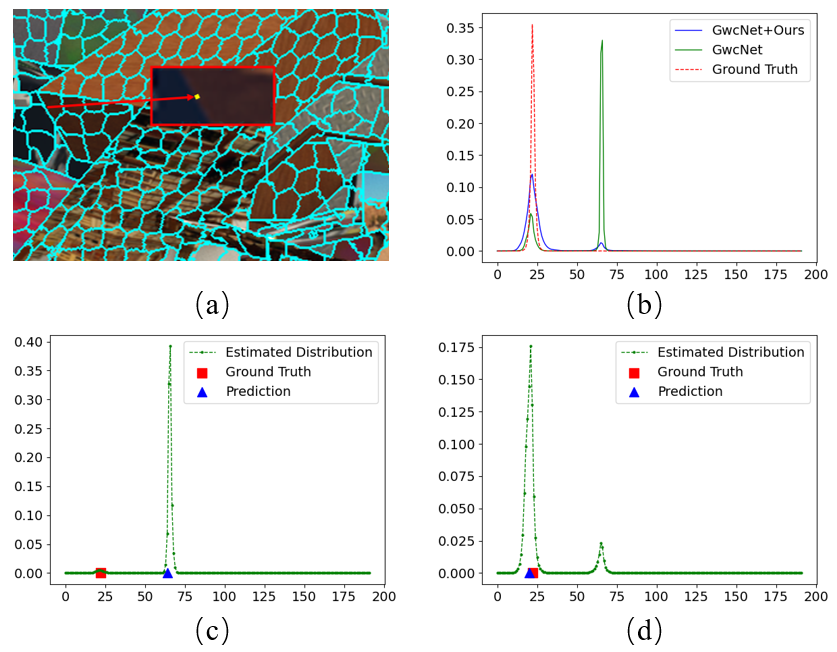}}
\caption{Visualization of the real output distribution at boundaries on Scene Flow dataset. (a) is the input image, and its partial enlargement. (b) represents the disparity probability distribution of the superpixel belonging to the brown region. (c) and (d) show the output probability distributions of a given pixel from GwcNet and GwcNet$+$Ours. Our proposed methods rectify the incorrect distributions and avoid smoothness bias. Please zoom in to see the details.}
\label{fig:motivation}
\end{figure}



\section{Introduction}
Stereo matching endeavors to establish dense correspondences between rectified stereo pairs, enabling the recovery of scene depth through triangulation\cite{SGM}. This technique finds broad applications in diverse fields, including robot navigation, augmented reality, and autonomous driving.

Recently, stereo models have demonstrated exceptional performance through the utilization of a cost volume-based architecture\cite{GC,PSM,GWC}, typically comprising four key steps: feature extraction, cost volume construction, cost aggregation, and disparity regression. Among these steps, cost aggregation stands out as the most crucial module, responsible for selecting the optimal match from numerous potential pairs and generating probability representations for the cost volume. However, state-of-the-art models face challenges in effectively addressing local ambiguities at boundaries, where definitively determining the pixel's belonging region is complex. This frequently leads to a multi-peaked distribution in the aggregated probability volume, giving rise to the problem of over-smoothing\cite{SMD,PSM+}.

In this study, we endeavor to rectify this mismatch and eliminate redundant information by incorporating a pixel relationship prior. Drawing inspiration from the premise that depth transitions smoothly within homologous regions\cite{Segment-Based_Stereo,ACV,PMSC}, we posit that depth discontinuities solely manifest between distinct regions. Hence, we introduce the concept of superpixels\cite{SLIC}, defined as clusters of contiguous and perceptually coherent pixels, offering a more coarse-grained representation of the image. Several recent superpixel segmentation methods have successfully integrated into various low-level tasks, including optical flow estimation, monocular depth estimation\cite{SPDE}, and depth completion, etc. They play a crucial role in decreasing the number of primitives in image processing, extracting similar features, and capturing image structure information.

Capitalizing on their inherent clustering and boundary properties, we integrate superpixel segmentation to produce a superpixel-level probability volume. Furthermore, the effectiveness of a strong-constraint disparity filtering strategy is limited due to the coarse-grained nature of superpixel representation, which cannot refine to each disparity level. To address this limitation, we model the ground truth at the superpixel level using a Laplace distribution\cite{SED} and apply cross-entropy loss to this representation, to suppress the multi-peaked issue during the cost aggregation into probability.
This superpixel training head proves highly effective in aiding aggregation, generating a more accurate probability representation for the cost volume, while simultaneously avoiding the need for additional computations and parameters during the inference stage of such resource-constrained tasks.
And we conducted experiments to explore its efficacy, as illustrated in Figure \ref{fig:motivation}, this approach facilitates the convergence of the probability volume within the same superpixel, rectifying outliers through the overall distribution.
To maintain color and spatial consistency, we adjust the sub-network's task orientation towards disparity reconstruction, leveraging the principle that pixels within superpixel blocks from color images share similar disparities. This enables attention weights, derived from the sub-network's semantic features, to effectively enhance local geometric consistency within the cost volume in the channel dimension, thereby encoding meaningful relationships between pixels.

\begin{figure}[t]
\begin{minipage}[c]{1.0\linewidth}
  \centering
  \centerline{\includegraphics[width=1.0\columnwidth]{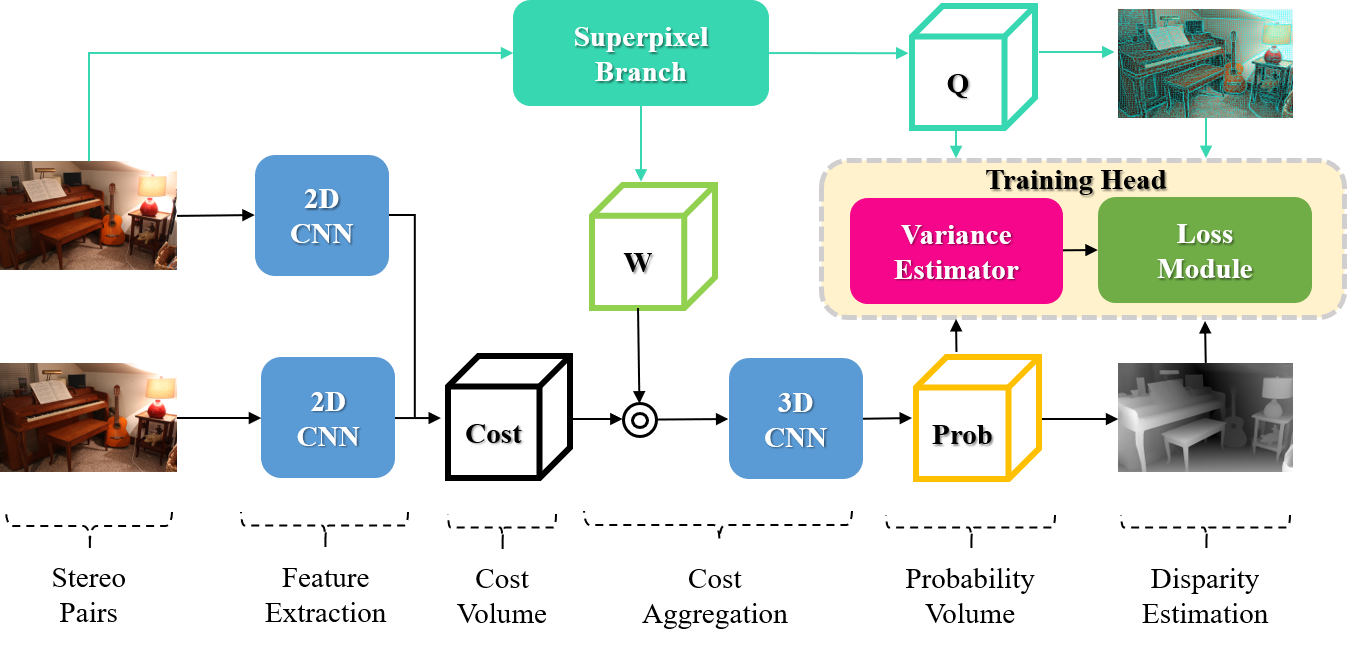}}
\end{minipage}

\caption{The proposed stereo matching framework consists of a stereo matching pipeline and a sub-network for superpixel segmentation. The superpixel branch (cyan) takes the left image as input and assists the stereo branch (black).}
\label{fig:framework}
\end{figure}

\section{Related Work}
The cost volume-based architecture is designed to enhance the accuracy of depth estimation by constructing and optimizing the cost associated with candidate disparitie. This volume is formed by concatenating or correlating feature maps extracted from the left and right images at various disparity levels. GCNet\cite{GC} pioneered the integration of a 3D encoder-decoder structure, utilizing soft-argmin-based disparity regression derived from a probabilistic cost volume. Subsequently, advancements such as the grouped-wise correlation cost volume introduced by GwcNet\cite{GWC} and the attention-based cost volume proposed by ACVNet\cite{ACV} aimed to augment the representational capacity of the cost volume. These end-to-end deep learning methods primarily supervise the disparity outputs, neglecting the rationality of their distributions. AcfNet\cite{ACF} addresses this issue by directly supervising the cost volume with unimodal ground truth distributions. However, due to its reliance on pixel-level operations, the network may struggle to learn scene structural information and could potentially overfit to a single dataset. 

Superpixels play a crucial role in local optimization and global consistency in stereo matching. Previous studies \cite{PMSC,SuperpixelAlphaExpansion} demonstrate that $\alpha$-expansion, which segments images into larger regions and assumes similar 3D plane labels within each segment, effectively optimizes disparity estimation by propagating consistent plane labels.
SFCN\cite{SFCN} effectively preserves object boundaries and fine-grained details by incorporating superpixels, which replaces conventional upsampling methods in the downsampling or upsampling scheme. However, this technique does not contribute to the matching process. In contrast, our approach delves deeper into the pixel relationship information inherent in the cost volume, emphasizing the collective impact of neighboring pixels on disparity estimation.

\vspace{-10pt}
\begin{figure}[htbp]
\begin{minipage}[c]{1.0\linewidth}
  \centering
  \centerline{\includegraphics[width=.8\columnwidth]{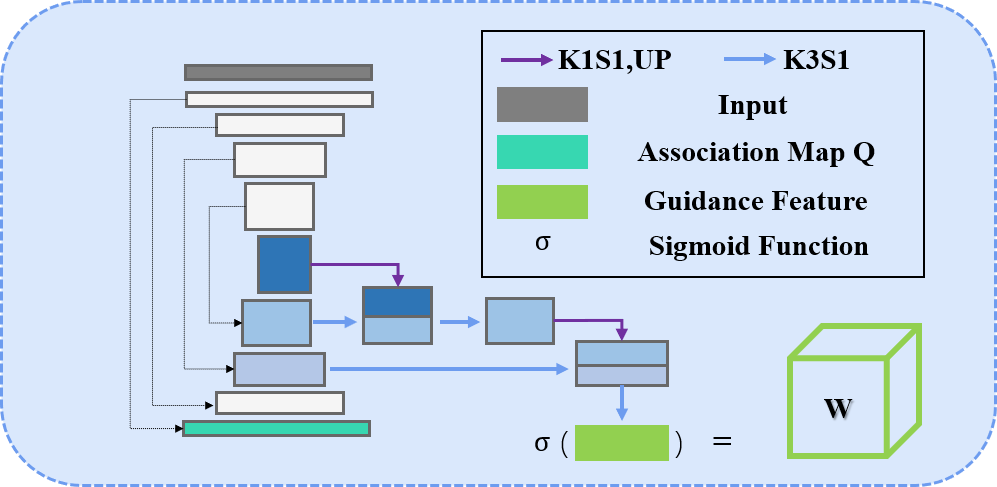}}
\end{minipage}
\caption{Superpixel guided channel excitation module.
The multi-scale short connections achieved through 2D convolutional kernels of varying sizes and strides combined with upsampling result in rich object context within the superpixel branch.}
\label{fig:channel excitation}
\end{figure}
\vspace{-20pt}

\section{Methods}
\subsection{Superpixel Guided Channel Excitation}
As shown in Figure \ref{fig:channel excitation}, superpixel segmentation is implemented using a standard encoder-decoder architecture with skip connections\cite{SFCN}. 
We argue that object context is crucial for accurate segmentation, as its multi-scale features contain valuable information about object shape and affinity. Therefore, we use the channel excitation to embed the object context into the cost volume.
Different from the CoEx\cite{COEX} method, where only involves excitation of the corresponding scaled cost volume features, we instead fuse hierarchical scales features $\phi(I_l)_k \in \mathbb{R}^{N \times \frac{H}{k} \times \frac{W}{k}}, k \in \{4, 8, 16\}$ from the sub-network. Through a simple multi-scale short connections denoted as $g$, we obtain the superpixel semantic guidance.
Before each cost aggregation, the guided cost volume excitation is calculated as:
\begin{equation}
\label{eq:1}
	\begin{split}
        C^{\prime}_{cost} &= \sigma(g(\phi(I_l)_k)) \odot C_{cost}
	\end{split}
\end{equation}
where $\sigma$ denotes the sigmoid function that converts the guidance into an attention weight map. These attention weights \textbf{W} emphasize both local consistency and discontinuity within the cost volume along the channel dimension.
And $\odot$ represents the Hadamard product after broadcasting the attention across the disparity dimension.
$\textbf{C}_{\textbf{cost}}$ (with the size of $N \times \frac{1}{4}D \times \frac{1}{4}H \times \frac{1}{4}W $) represents the 4D cost volume constructed from the features of the left and right images. 
This process generates a geometrically encoded cost volume, $\textbf{C}^{\prime}_{\textbf{cost}}$, which allows 3D convolutions to aggregate information from neighboring pixels and capture geometric relationships inherent in the data.

\vspace{-10pt}
\begin{figure}[htbp]
\begin{minipage}[c]{1.0\linewidth}
  \centering
  \centerline{\includegraphics[width=1.0\columnwidth]{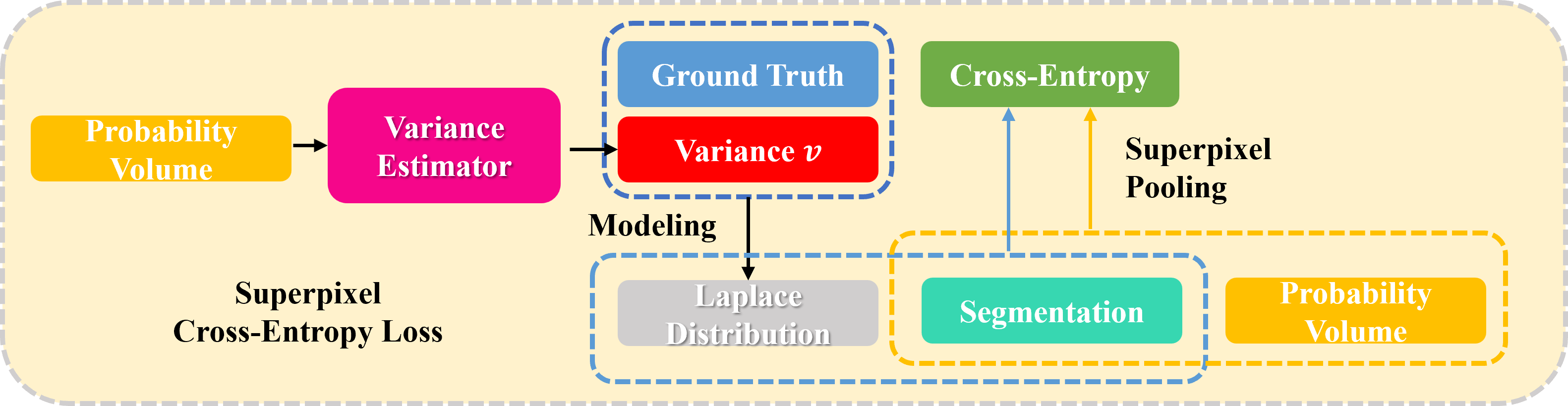}}
\end{minipage}

\caption{The main components of the joint learning training head, which combines the output results from two branches, consist of a variance estimator for predicting matchbility and a superpixel pooling module, all driven by the cross-entropy loss function.}
\label{fig:training head}
\end{figure}

\vspace{-20pt}
\subsection{Superpixel Pooling of Probability Volume}

\textbf{Laplace Distribution.} In an ideal scenario, the disparity probability distribution manifests itself in a unimodal form, where the probability values diminish with the distance from the true matching pixel, peaking at the ground truth disparity value. To more accurately depict the variance in disparity probability distributions across different matching regions, we adaptively model a unimodal distribution akin to AcfNet\cite{ACF} as follows: 
\begin{equation}
    P^{gt}(d) = \text{softmax}\left(- \frac{|d - d^{gt}|}{v}\right)
\end{equation}
As depicted in Figure \ref{fig:training head}, the variance $v$ is computed based on the aggregated cost volume (i.e. probability volume). Challenging pixels often exhibit multi-modal probability distributions, with their variance typically being large. 
 This parameter controls the sharpness of the peak around the true disparity, adjusting it according to the matchbility. Specifically, when a point struggles to distinctly delineate the pixel's region of belonging or resides within a region characterized by weak textural attributes during stereo matching, it exhibits a comparatively smoother peak. This adaptive modeling enhances the precision of disparaty estimation across various matching scenarios.

\noindent\textbf{Superpixel Pooling.}
Given the predicted superpixel association probability map $\textbf{Q} \in \mathbb{R}^{{\left\vert N_s \right\vert} \times H\times W}$ for image $I_l$, where $N_s$ represents the 9 sets of surrounding initial grid cells associated with each pixel $p$, we obtain the superpixel label map $m$ by assigning each pixel to its most likely superpixel using $m = {{\arg\max}\, Q(p)}$. And its inverse mapping $\tilde{m}$, which represents the pixel index of each superpixel label.

To capture the disparity probability distribution within each superpixel, we leverage $m$ and the aggregated volume $\textbf{C}_{\textbf{prob}}$ (with the size of $D \times H \times W $) to generate a superpixel probability volume $P_s$, defined as:

\begin{equation}
\label{eq:2}
P_s = \left(\prod_{p\in \tilde{m}_s} C_{prob}(p)\right)^{\frac{1}{n}} 
\end{equation}
where $n$ denotes the number of pixels within the specific superpixel $s$.
To ensure numerical stability, circumvent underflow issues arising from probabilistic multiplication, and simplify computational complexity, we conduct the pooling process in logarithmic space:
\begin{equation}
\label{eq:3}
\ln(P_s) = \frac{1}{n} \sum_{p\in \tilde{m}_s} \ln(C_{prob}(p))
\end{equation}

To recover the original superpixel probability volume from logarithmic space, we apply exponential operations. 
We can perform superpixel geometric mean pooling over the modeled ground truth from the preceding section or probability volume, to obtain a superpixel-level probability representation. The probability distributions of pixels exhibit a collective influence, where their interactions shape the overall probability distribution within superpixels. Notably, this superpixel probability volume is generated solely for supervision during training, ensuring no added computational or memory demands during inference. 

\subsection{Training Head}
\textbf{Loss for Single Tasks.}
After getting the final probability volume, the soft-argmin operation is used to compute disparity for each pixel by taking the expected value\cite{GC}. To ensure regression focuses on the most probable mode, we utilize the top $k$ values from the probability volume:
\begin{equation}
\label{eq:4}
    \hat{d} = \sum_{d \in \{d_1,d_2,...,d_k\}} d \times Softmax(C_{prob}(d))
\end{equation}

The output results of the two branches as shown in Figure \ref{fig:framework} are fed into the training head for final supervision. For the disparity estimation task, we mainly use Smooth $L_1$ Loss, which has been widely used in various regression tasks:
\begin{equation}
\label{eq:5}
    \mathcal{L}_{regression} = \frac{1}{N}\sum_{p}Smooth_{L_1}(d_p,\hat{d}_p)
\end{equation}
In equation \ref{eq:5}, $\hat{d}_p$ and $d_p$ are the predicted disparity and corresponding groundtruth respectively, $N$ is the number of valid pixels. During training, we supervise the estimation of each regression stage.

As for the superpixel segmentation auxiliary task, to encourage the segmentation network to generate superpixels that effectively represent disparity, we further define a disparity reconstruction loss \cite{SFCN} \cite{SSN}:
\begin{equation}
\label{eq:7}
    \mathcal{L}_{recon} = \frac{1}{N} \sum_{p}  \left\Vert d_p - d_p' \right\Vert_1 + w \cdot \left\Vert p - p' \right\Vert_2 
\end{equation}
where $d'$ and $p'$ represent the superpixel reconstruction results obtained by left multiplying association map $\tilde{Q} \hat{Q}^T$, the row and column-normalizd association map $Q$, and $w$ controls the compactness of the superpixel.

\noindent\textbf{Superpixel Cross-Entropy Loss.} The probability after adaptive unimodal distribution modeling and superpixel pooling incorporates the contributions of neighboring pixels, emphasizing similar distributions that reflect the dominant trend within a superpixel.
\begin{equation}
\label{eq:6}
    \mathcal{L}_{sce}= -\frac{1}{N_s}
    \sum^{D-1}_{d=0}P_{s}^{gt}(d)\cdot\log P_s(d)
\end{equation}
which measures the similarity between the prediction $P_s$ and the constructed ground truth $P_s^{gt}$.
The total loss function is the sum of these three components:
\begin{equation}
\label{eq:8}
    \mathcal{L}_{total} =  \mathcal{L}_{regression} + \lambda\mathcal{L}_{sce}  + \mu\mathcal{L}_{recon}
\end{equation}
During the training, We heuristically set $\lambda=1$ and $\mu=0.1$ in our experiments.

\section{Experiments}
\label{sec:pagestyle}
\subsection{Implementation Details}
We implemented the proposed method using PyTorch and conducted experiments on NVIDIA RTX 3090 GPUs, employing the Adam optimizer with $\beta_1 = 0.9$ and $\beta_2=0.999$. To facilitate model generalization, we augmented input images during the training phase by employing random cropping to a size of $H = 256$ and $W = 512$.

For the Scene Flow dataset, we trained GwcNet integrated with the proposed techniques for a total of 16 epochs. An initial learning rate of 0.001 was applied, strategically reduced by a factor of 2 after epochs 10, 12, and 14 to ensure model convergence. A batch size of 4 was used to optimize memory utilization. The weighting factor $w$ was set to $5\times10^{-3}$ for appropriate disparity reconstruction loss contribution and $k$ was set to 6 for the superior performance observed in our prior work.
To further enhance model performance, we fine-tuned the models pre-trained on Scene Flow using the KITTI and Middlebury datasets. This fine-tuning process involved 300 additional epochs with an initial learning rate of 0.001, reduced by a factor of 10 after 200 epochs to facilitate fine-grained adjustments in the later training stages.

To ensure consistency and focus within the defined disparity range, we excluded ground truth disparities falling outside the interval $[0, D_{max}]$ during experiments, where $D_{max}$ was set to 192.

\vspace{-10pt}
\begin{table}[htbp]
    \caption{Ablation study on Scene Flow finalpass dataset.}
    \centering
    \setlength{\tabcolsep}{1.5mm}{%
        \begin{tabular}{c|cccc|c|c|c|c}
            \hline
            Method & $\mathcal{L}_{ce}$ & $\mathcal{L}_{sce}$ & $\mathcal{L}_{reconC}$ & $\mathcal{L}_{reconD}$ & EPE (px) & 1 px (\%)  & 2 px (\%) & 3 px (\%) \\
            \hline\hline
            \multirow{2}{*}{GwcNet} & \text{-} & \text{-} & \text{-} & \text{-} &0.765 &8.03 &4.47 & 3.30 \\
            & \text{-} & \checkmark & \text{-} & \checkmark & 0.670 & 6.50 & 3.78 & 2.86 \\
            \hline
            \multirow{4}{*}{$+$ SGCE}
            & \text{-} & \text{-} & \checkmark & \text{-} & 0.645 & 6.60 & 3.71 & 2.74 \\
            & \text{-} & \text{-} & \text{-} & \checkmark & 0.626 & 6.44 & 3.63 & 2.70 \\
            & \checkmark & \text{-} & \text{-} & \checkmark & 0.622 & 6.49 & 3.65 & 2.71\\
            & \text{-} & \checkmark & \text{-} & \checkmark & \textbf{0.596} & \textbf{6.00} & \textbf{3.41} & \textbf{2.54} \\
            \hline
        \end{tabular}%
    }
    \label{tab:albation} 
\end{table}
\vspace{-20pt}

\subsection{Modules Designed}
To meticulously evaluate the contributions of individual components within our proposed methodology, we conducted a comprehensive ablation study on the Scene Flow dataset. GwcNet \cite{GWC} served as the baseline, and we systematically examined the effectiveness of Superpixel Guided Channel Excitation (SGCE), $\mathcal{L}_{sce}$, and $\mathcal{L}_{recon}$ by employing various experimental settings.

\begin{figure}[htbp]
\begin{minipage}[c]{1.0\linewidth}
  \centering
  \centerline{\includegraphics[width=1.0\columnwidth]{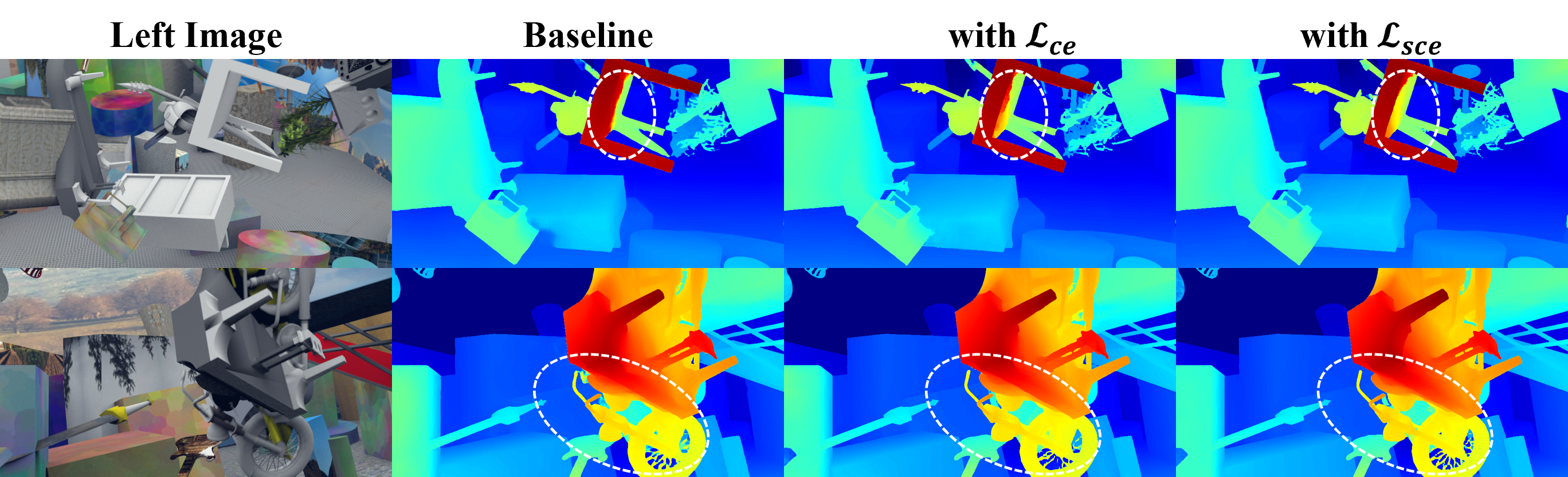}}
\end{minipage}
\caption{Qualitative comparisons of ablation study on Scene Flow test set.}
\label{fig:ablation sf}
\end{figure}

Initially, we focused on assessing the efficacy of the proposed loss function without introducing any structural modifications to the baseline network. Figure \ref{fig:ablation sf} illustrates the improved performance, particularly highlighting the enhancement in object boundary detailing, attributed to $\mathcal{L}_{sce}$.
Subsequently, we performed comparisons between superpixel cross entropy loss and regular cross entropy loss\cite{PSM+}, as well as investigations into the influence of depth- and color-based reconstruction losses.
The results, as presented in Table \ref{tab:albation}, offer compelling insights:
Regular loss functions, when employed in isolation, can potentially exert detrimental effects on performance. Our proposed loss components, in contrast, demonstrate consistent improvements across all evaluated stereo matching error metrics, surpassing the baseline results.

\begin{table}[htbp]
    \caption{Universality study on Scene Flow finalpass dataset.($*$ denotes the finalpass reproduced result)}
    \centering
    \setlength{\tabcolsep}{4mm}{%
        \begin{tabular}{l|c|c|c}
            \hline
            Method & EPE (px) & D1 (\%) & SEE (px)\\
            \hline
            PSMNet$^*$\cite{PSM} & 1.11 & \textbf{2.47} & 4.42 \\
            PSMNet-TH& \textbf{1.06} & 2.49 & \textbf{3.07} \\                   
            \hline
            MobileStereo\cite{MOBILESTEREO}& 1.14 & 4.40 & 4.41 \\
            MobileStereo-TH& \textbf{0.92} & \textbf{3.21} & \textbf{3.63} \\                   
            \hline
            PCWNet$^*$\cite{PCW}& 0.84 & 2.80 & 3.83 \\
            PCWNet-TH& \textbf{0.74} & \textbf{2.52} & \textbf{3.66} \\                   
            \hline
        \end{tabular}%
    }
    \label{tab:universality}
\end{table}
\vspace{-10pt}

\subsection{Universality of the Training Head}
To demonstrate the universality of our proposed training head, we seamlessly integrate it into three state-of-the-art models, namely PSMNet\cite{PSM}, MobileStereo\cite{MOBILESTEREO} and PCWNet\cite{PCW}. We then compare the performance of the original models with the integrated versions, denoted as PSMNet-TH, MobileStereo-TH, and PCWNet-TH, respectively. The evaluation, as presented in Table III, includes a dedicated metric for quantifying the quality of disparities at boundaries, referred to as SEE (Soft Edge Error).
It is important to note that we have not validated the universality and effectiveness of our approach on iterative refinement architectures, such as RAFT-Stereo\cite{RAFT}. This is due to the fact that our loss function is tailored to optimize the probabilistic form of the cost volume.

\vspace{-10pt}
\begin{table}[htbp]
    \caption{Quantitative evaluation on Scene Flow test set with the popular approaches.}
    \centering
    \resizebox{\linewidth}{!}{%
        \begin{tabular}{cccccccc}
            \hline	
            Method & PSMNet\cite{PSM} & GwcNet\cite{GWC} & SSPCV-Net\cite{SSPCV} & EdgeStereo\cite{EDGESTEREO} 
            & AcfNet\cite{ACF} & ACVNet\cite{ACV} & GwcNet$+$Ours  \\
            \hline
            EPE (px) & 1.09 & 0.76 & 0.87 & 1.11 
            & 0.86 & \textbf{0.48} & \underline{0.59} \\
            \hline
        \end{tabular}%
    }
    \label{tab:scene_flow_benchmark}
    \\{\smallskip}
    \textbf{Bold}: Best, \underline{Underline}: Secondary
\end{table}
\vspace{-20pt}

\subsection{Performance Evaluation}
\label{sec:3.3}
\textbf{Scene Flow Dataset}.
To assess model performance in real-world indoor scenes, we utilized the Middlebury dataset, consisting of 15 training image pairs and 15 test pairs. Experiments were conducted using half-resolution images to align with dataset conventions.
Table \ref{tab:scene_flow_benchmark} showcases the outstanding performance of our approach. Notably, it ranks second among all competing algorithms, achieving a remarkable 22\% reduction in EPE when integrated with GwcNet. These results emphatically demonstrate the effectiveness of our methodology in enhancing disparity estimation accuracy. 

\begin{figure}[htbp]
\begin{minipage}[c]{1\linewidth}
  \centering
  \centerline{\includegraphics[width=\columnwidth]{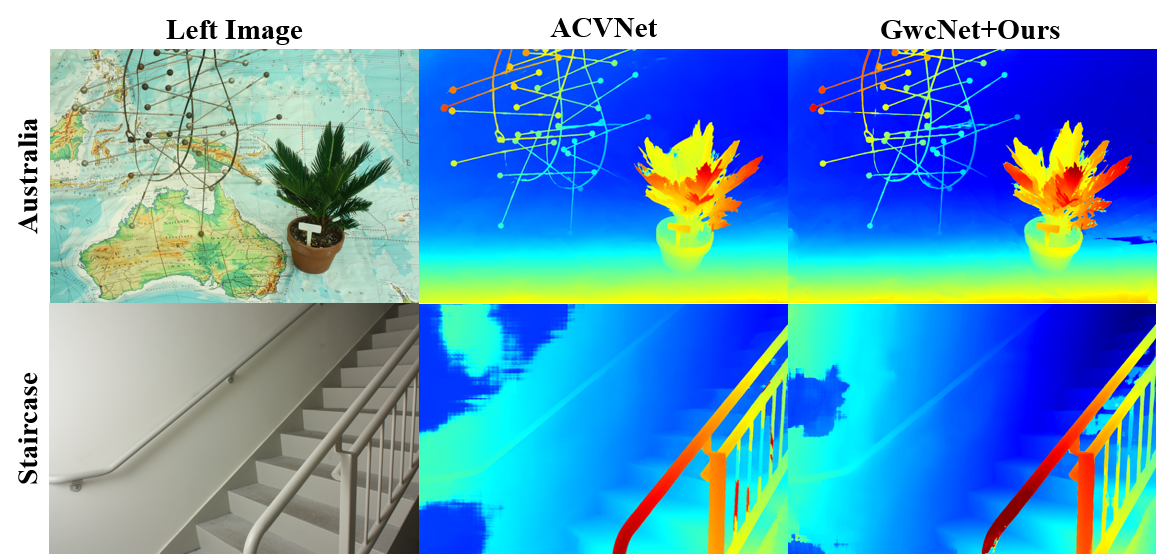}}
\end{minipage}
\caption{Qualitative results on the Middlebury test set compared to the top end-to-end deep learning approach ACVNet\cite{ACV}.}
\label{fig:mid}
\end{figure}
\vspace{-10pt}

\noindent\textbf{Middlebury Dataset}.
To assess model performance in real-world indoor scenes, we utilized the Middlebury dataset. Figure \ref{fig:mid} visually compares the disparity quality of our approach against other leading method on the test dense leaderboard. The results reveal several distinct advantages: 
sharper transitions at object boundaries, indicating enhanced edge preservation and detail capture; consistent disparity predictions within individual objects, demonstrating robust depth estimation.

\begin{figure}[htbp]
\begin{minipage}[c]{1.0\linewidth}
  \centering
  \centerline{\includegraphics[width=\columnwidth]{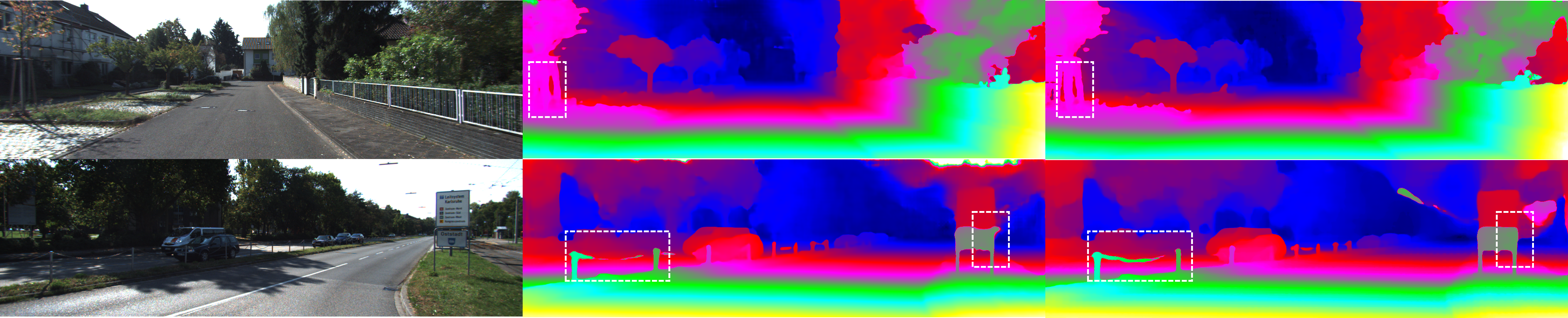}}
\end{minipage}

\caption{Qualitative results on the KITTI 2012 (top) and KITTI 2015 (bottom) test set. White box highlighted the improvement of details.}
\label{fig:kitti}
\end{figure}
\vspace{-10pt}

\noindent\textbf{KITTI}.
To evaluate model performance in real-world driving scenarios, we employed the KITTI 2015 and KITTI 2012 datasets, both capturing challenging outdoor scenes. KITTI 2015 offers 200 training stereo image pairs with sparse ground-truth disparities and 200 testing pairs without ground truth, while KITTI 2012 provides 194 training pairs and 195 testing pairs.
As presented in Tables \ref{tab:kitti12} and \ref{tab:kitti15}, our approach demonstrates competitive performance, aligning with the results of leading networks in the field. Due to the sparse ground truth in the dataset, performance degradation occurs during fine-tuning of superpixel branches. Additionally, in large scenes, segmentation areas may slightly deviate from our principle of disparity consistency. These challenges indicate the potential of our proposed methods for further enhancement when dealing with complex scenes.
AcfNet \cite{ACF}, while effective, relies on pixel-level uncertainty supervision and unimodal distribution modeling, potentially limiting its ability to fully leverage contextual information from neighboring pixels. Our approach, in contrast, explicitly addresses this limitation through superpixel-based guidance, resulting in superior performance.
Furthermore, comparisons with SSPCV-Net\cite{SSPCV} and EdgeStereo\cite{EDGESTEREO} highlight the advantages of superpixels. Unlike these methods, which introduce subnetworks for segmentation or edge detection, our superpixel-based approach implicitly considers both semantic classes and boundary information, leading to more comprehensive guidance for stereo matching.

\begin{table}[htbp]
        \caption{Quantitative evaluation on KITTI 2012 test set.}
        \label{tab:kitti12} 
	\centering 
    \setlength{\tabcolsep}{4.5mm}{%
	\begin{tabular}{lccc cccc}
		\hline
		\multirow{2}{*}{Method} & \multicolumn{2}{c}{3px (\%)} & \multicolumn{2}{c}{5px (\%)} & \multicolumn{2}{c}{EPE (px)}\\
                                    &noc &all
                                    &noc &all 
                                    &noc &all \\
		\hline
		SSPCV-Net\cite{SSPCV}& 1.47 & 1.90 & 0.87 & 1.14 & 0.5 & 0.6 \\
            EdgeStereo-V2\cite{EDGESTEREO}& 1.46 & 1.83 & 0.83 & 1.04 & 0.4 & 0.5 \\
            CoEx\cite{COEX}& 1.55 & 1.93 & 0.91 & 1.13 & 0.5 & 0.5 \\
            AcfNet\cite{ACF}& 1.17 & 1.54 & 0.77 & 1.01 & 0.5 & 0.5\\
            RAFT-Stereo\cite{RAFT}& 1.30 & 1.66 & 0.86 & 1.11 & 0.4 & 0.5\\
            ACVNet\cite{ACV}&\underline{1.13}&\underline{1.47}&\textbf{0.71}&\textbf{0.91}&0.4&0.5\\
            IGEV-Stereo\cite{IGEV}& \textbf{1.12} & \textbf{1.44} & 0.73 & 0.94 & \textbf{0.4} & \textbf{0.4}\\
            \hline
            GwcNet-gc\cite{GWC}& 1.32 & 1.70 & 0.80 & 1.03 & 0.5 & 0.5\\
            GwcNet$+$Ours& 1.18 & 1.50 & \underline{0.72} &\underline{0.93} & \textbf{0.4} & \underline{0.5}\\
            \hline
	\end{tabular}%
 }
\end{table}
\vspace{-25pt}
\begin{table}[htbp]
        \caption{Quantitative evaluation on KITTI 2015 test set.}
        \label{tab:kitti15} 
	\centering 
    \setlength{\tabcolsep}{4mm}{%
	\begin{tabular}{lccc cccc}
		\hline
		\multirow{2}{*}{Method} & \multicolumn{3}{c}{NOC (\%)} & \multicolumn{3}{c}{ALL (\%)} \\
                                    &bg &fg &all
                                    &bg &fg &all \\
		\hline
		SSPCV-Net\cite{SSPCV}&1.61&3.40&1.91&1.75&3.89&2.11\\
            DeepPruner-Best\cite{DEEPPRUNER}&1.71&3.18&1.95&1.87&3.56&2.15\\
            EdgeStereo\cite{EDGESTEREO}&1.72&3.41&2.00&1.87&3.61&2.16\\
            CoEx\cite{COEX}&1.62&\underline{3.09}&1.86&1.74&\underline{3.41}&2.02\\
            ACVNet\cite{ACV}&\textbf{1.37}&\textbf{3.07}&\textbf{1.65}&\textbf{1.26}&\textbf{2.84}&\textbf{1.52}\\
            \hline
            GwcNet-g\cite{GWC}& 1.61&3.49&1.92&1.74&3.93&2.11\\ GwcNet$+$Ours&\underline{1.48}&3.20&\underline{1.76}&\underline{1.60}&3.59&\underline{1.93}\\
            \hline
	\end{tabular}%
 }
\end{table}

\vspace{-20pt}
\section{Conclusion}
\label{sec:typestyle}

In this paper, we propose a novel stereo matching approach that combines superpixels and cross-entropy loss, resulting in enhanced accuracy and robustness. Our method utilizes a superpixel probability volume to enable effective learning of regional features and outlier correction. Through seamless integration with classical stereo matching networks, our approach demonstrates significant improvements across various datasets. We anticipate its potential benefits for downstream tasks, such as stereo-based 3D reconstruction.
%
%
%
%

\end{document}